# Evaluating influence diagrams with decision circuits

**Debarun Bhattacharjya and Ross D. Shachter**
Department of Management Science and Engineering
Stanford University
Stanford, CA 94305, USA
E-mail: debarunb@stanford.edu, shachter@stanford.edu

## Abstract

Although a number of related algorithms have been developed to evaluate influence diagrams, exploiting the conditional independence in the diagram, the exact solution has remained intractable for many important problems. In this paper we introduce decision circuits as a means to exploit the local structure usually found in decision problems and to improve the performance of influence diagram analysis. This work builds on the probabilistic inference algorithms using arithmetic circuits to represent Bayesian belief networks [Darwiche, 2003]. Once compiled, these arithmetic circuits efficiently evaluate probabilistic queries on the belief network, and methods have been developed to exploit both the global and local structure of the network. We show that decision circuits can be constructed in a similar fashion and promise similar benefits.

## 1 INTRODUCTION

Influence diagrams were introduced as a representation to facilitate communication in practical decision problems [Howard and Matheson, 1984]. They have also proven to be powerful computational aids. Exact methods for evaluating influence diagrams either work directly on the influence diagram [Shachter, 1986], or on related graphical structures such as valuation networks [Shenoy, 1992] or junction trees [Shachter and Peot, 1992; Jensen et al, 1994].

Advances in Bayesian belief network computation have often led to new approaches to evaluate influence diagrams. Earlier research work has shown the correspondence between probabilistic inference algorithms and methods for evaluating influence diagrams [Cooper, 1988; Shachter, 1988; Shachter and Peot, 1992]. Darwiche's (2003) recent work on arithmetic circuits has demonstrated the power of using both local structure (in the form of determinism and context-specific independence) and global structure (conditional independence) for probabilistic inference in belief networks. In this paper, we show how the concepts of arithmetic circuits can be used to evaluate influence diagrams efficiently, either directly or in a modified form we call decision circuits.

Inference on belief networks using arithmetic circuits has been shown to perform at least as well as the state-of-the-art jointree methods [Park and Darwiche, 2004]. The challenge is to construct even better circuits than those corresponding to jointrees. Some techniques that use local structure have recently been devised [Darwiche 2002a; Chavira and Darwiche, 2005] and experimental results show that compact circuits can be constructed for belief networks with high treewidth and sufficient local structure [Chavira and Darwiche, 2005; Chavira et al, 2006]. These methods are based on propositional theories and conversion from one logical form to another [Darwiche, 2001; Darwiche 2002b].

In this paper, we show how decision circuits can be constructed and evaluated. Decision circuits are constrained by the order restrictions dictated by which uncertainties have been observed at the time decisions are made. Nonetheless, decision problems in practice have considerable local structure from asymmetry and conditional alternatives, and previous research has shown how exploiting local structure can improve efficiency [Gomez and Cano, 2003]. The recent advances in arithmetic circuits could be used to evaluate practical decision problems for which traditional approaches have proven intractable.

In section 2, we review the concepts of influence diagrams using some simple examples. We briefly summarize some of the previous literature on arithmetic circuits in section 3. We demonstrate how all influence diagrams can be solved directly using arithmetic circuits in section 4. Decision circuits are introduced in section 5. We show how to construct decision circuits from variable elimination and establish that they can be created in time and space of the same complexity as existing state-of-the-art methods. We present methods for finding the optimal policy and the maximal expected utility. Finally, section 6 describes our conclusions and directions for future work.

## 2 INFLUENCE DIAGRAMS

In this section we introduce the concepts and notation of belief networks and influence diagrams that we will use throughout the paper. These models represent the beliefs of a single rational decision maker.

Both *Bayesian belief networks* and *influence diagrams* are directed acyclic graphical models for reasoning under uncertainty. All of the nodes in a belief network correspond to uncertain variables, while influence diagrams also include decisions and values. Variables are denoted by upper-case letters ($X$) and their values by lower-case letters ($x$). A bold-faced letter indicates a set of variables. We will refer interchangeably to a node in the diagram and the corresponding variable. We will draw uncertainties as ovals, decisions as rectangles and values as diamonds. The parents of an uncertain or value variable condition its probability distribution. If $X$ is a variable with parents $\mathbf{U}$, then $X\mathbf{U} = X \cup \mathbf{U}$ is called the family for variable $X$. On the other hand, the parents of a decision variable are those variables whose values will be observed before the decision must be made.

Consider the three examples shown in Figure 1. Figure 1a presents a belief network with two nodes, labelled W (Weather) and B (Bring umbrella). We are interested in knowing whether a friend will bring an umbrella, and our belief can be different if we observe the weather. If there were no arc between nodes W and B, then we would be asserting that observing the weather would not change our belief about the umbrella. Figure 1b shows the influence diagram for a simple decision problem in which our friend chooses whether to bring an umbrella based on her belief about the weather and her preferences, represented by the node U (Utility). She ranks her preferences for all possible prospects of weather and umbrella using a von Neumann-Morgenstern utility function [von Neumann and Morgenstern, 1947], where having $U = 1$ is at least as good as anything that can happen, and having $U = 0$ is at least as bad. Thus, she can represent her preferences by the probability that $U = 1$ for each prospect. Figure 1c shows the influence diagram for a more complicated decision problem, where she can Gather evidence (G), such as to purchase a weather Report (R). Her information gathering decision and the report will be known to her when she decides whether to bring her umbrella. We re-visit these examples in later sections.

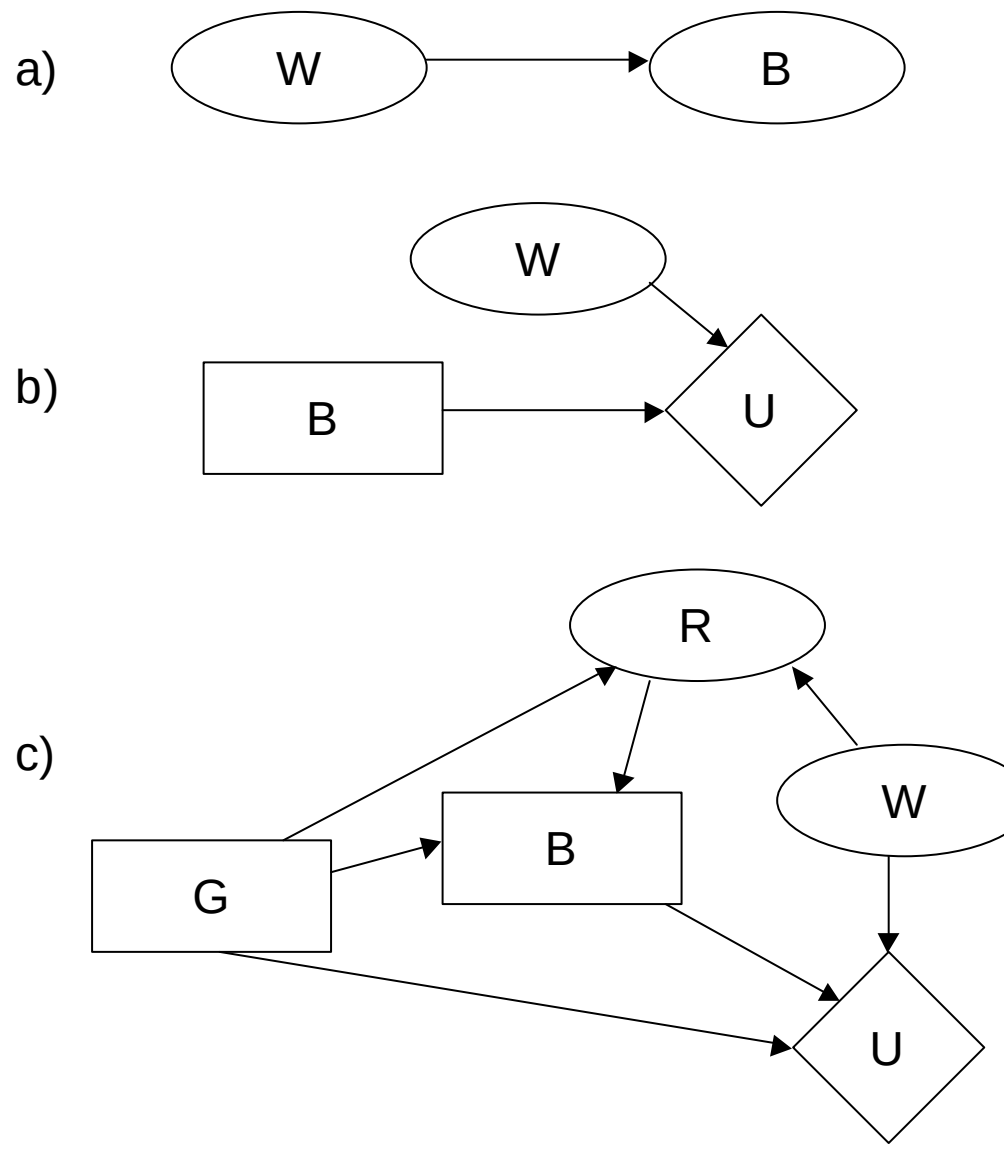

Figure 1: Some influence diagram examples.

When decisions are represented by separate nodes, as shown in Figure 1c, the diagram is said to be in *extensive form*. Alternatively, when there is only one decision node and it has no parents, as shown in Figure 1b, the diagram is said to be in *normal form* [Savage, 1954; Raiffa, 1968]. In fact, we can convert any diagram in extensive form to one in normal form by considering different *strategies*, one for each possible combination of observed uncertainties and decision alternatives. Therefore, the number of strategies can be very large, and it is often much more efficient to maintain the problem in extensive form.

The choice of Strategy (S) makes all of the decisions deterministic functions of the uncertainties which will be observed beforehand. For example, the diagram shown in Figure 2 represents the normal form of the diagram shown in Figure 1c. There is a single strategy decision, while the other decisions become deterministic functions (shown as double ovals) of the strategy and the observed uncertainties. Each strategy includes a choice for gathering information and a choice for bringing umbrella for each possible value of weather report.

Although influence diagrams with multiple additive or multiplicative value nodes can provide some additional computational savings [Tatman and Shachter, 1990], we assume there is a single value node for readability

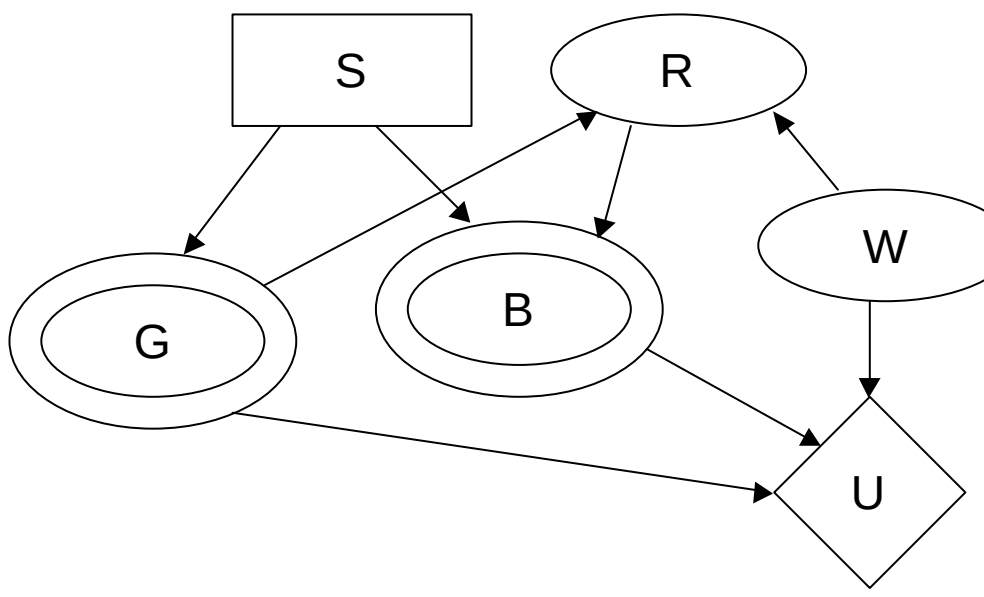

Figure 2: Normal form for the influence diagram shown in Figure 1c.

and simplicity.

The general problem we are solving is to determine the decisions $\mathbf{D}$ that maximize $P(U = 1)$ when the variables $\mathbf{E}$ have already been observed and the parents of each decision $D$ are observed before it is made. It is inconsistent to make observations that tell us anything about the decisions we are yet to make, so we require that $\mathbf{E}$ not be responsive in any way to $\mathbf{D}$ [Heckerman and Shachter, 1995]. This is enforced by not allowing the nodes in $\mathbf{E}$ to be descendants of the nodes in $\mathbf{D}$. Another inconsistency arises if information known before one decision is made is not known before subsequent decisions. To be a valid influence diagram for a rational individual there must exist at least one "no forgetting" ordering of the decisions such that all earlier observations and decisions are observed before any later decisions are made.

## 3 ARITHMETIC CIRCUITS

We explain basic concepts regarding arithmetic circuits [Darwiche, 2003]; in section 5 we will introduce decision circuits as an extension.

Belief networks are associated with a unique multi-linear function over two kinds of variables, *evidence indicators* and *network parameters*. An evidence indicator $\lambda_x$ is associated with each possible instantation $x$ of each network variable $X$. Similarly, there is a network parameter $\theta_{x|\mathbf{u}}$ for each possible instantation $x\mathbf{u}$ of family $X\mathbf{U}$. The network parameters represent the conditional probabilities, $\theta_{x|\mathbf{u}} = P(x|\mathbf{u})$, and the evidence indicators are binary (0-1) variables, with $\lambda_x = 0$ whenever $X$ has been observed taking another value. Each *term* in the multi-linear function corresponds to one instantiation $\mathbf{x}$ of all the network variables $\mathbf{X}$, and is the product of all evidence indicators and network parameters consistent with $\mathbf{x}$. In general, the multi-linear function for a belief network is $f = \sum_{\mathbf{x}} \prod_{x\mathbf{u} \sim \mathbf{x}} \lambda_x \theta_{x|\mathbf{u}}$ where the sum is over every instantiation of all of the variables in the network and $x\mathbf{u} \sim \mathbf{x}$ represents all families consistent with $\mathbf{x}$.

Consider the belief network of Figure 1a. Suppose there are two possible states for the weather, $w$ and $\bar{w}$, and two states for whether our friend brings an umbrella, $b$ and $\bar{b}$. The multi-linear function for this network is $f = \lambda_w \lambda_b \theta_w \theta_{b|w} + \lambda_{\bar{w}} \lambda_b \theta_{\bar{w}} \theta_{b|\bar{w}} + \lambda_w \lambda_{\bar{b}} \theta_w \theta_{\bar{b}|w} + \lambda_{\bar{w}} \lambda_{\bar{b}} \theta_{\bar{w}} \theta_{\bar{b}|\bar{w}}$.

This function has several special properties for answering inference queries. By setting the evidence indicators to 0 or 1, we can find the probability of observing any set of network variables $\mathbf{E}$. For instance, if we assign evidence to be $\mathbf{e} = \bar{b}$ by setting $\lambda_b = 0$ and all the other three evidence indicators as 1, the function returns $P(\bar{b}) = \theta_w \theta_{\bar{b}|w} + \theta_{\bar{w}} \theta_{\bar{b}|\bar{w}}$. The general procedure involves setting all of the evidence indicators to be consistent with the evidence $\mathbf{e}$ to compute $P(\mathbf{e})$.

Furthermore, the partial derivatives of the multi-linear function are also related to common probabilistic inference queries. Suppose $\mathbf{e}$ is evidence on network variables $\mathbf{E}$ and $X$ is any random variable ($X$ may or may not be in $\mathbf{E}$). $\mathbf{e} - X$ signifies the evidence such that any instantiation of $X$ is retracted. We list lemmas with some important results on partial derivatives as proven in Darwiche (2003).

**Lemma 1.** *For every variable $X$ and evidence $\mathbf{e}$, we have: $P(x, \mathbf{e} - X) = \frac{\partial f}{\partial \lambda_x}(\mathbf{e})$.*

**Lemma 2.** *For every variable $X$ and evidence $\mathbf{e}$, we have: $P(\mathbf{e} - X) = \sum_x \frac{\partial f}{\partial \lambda_x}(\mathbf{e})$.*

**Lemma 3.** *For every family $X\mathbf{U}$ and evidence $\mathbf{e}$, we have: $P(x, \mathbf{u}, \mathbf{e}) = \theta_{x|\mathbf{u}} \frac{\partial f}{\partial \theta_{x|\mathbf{u}}}(\mathbf{e})$.*

In general, there are an exponential number of terms in the multi-linear function associated with a belief network. Nonetheless, it can be efficiently represented, evaluated, and differentiated using arithmetic circuits.

**Definition 1.** *An arithmetic circuit is a rooted, directed acyclic graph whose leaf nodes are constants or variables and all other nodes represent either summation or multiplication. The size of an arithmetic circuit is the number of edges it contains.*

The arithmetic circuit for our simple belief network from Figure 1a is shown in Figure 3. Note that the value of the multi-linear function is computed at the root, hence an *upward pass*, starting from the leaves and ending at the root, will compute $f(\mathbf{e})$, where $f(\mathbf{e}) = P(\mathbf{e})$. This process is also known as *evaluating* the circuit.

We can calculate partial derivatives through a subsequent *downward pass*, in which the parents are visited before the children. This is known as *differentiating* the circuit, and is a bit more involved than the up-ward pass. Let $v$ be an arbitrary node in the circuit and also its value (determined in the upward sweep). Similarly, $f$ denotes the root node and also its value, the output of the circuit. We are interested in the partial derivative $\partial f/\partial v$ for all nodes $v$, recognizing that $f$ is a function of all of the node values. If $v$ is the root node $f$, then $\partial f/\partial v = 1$. For any other node $v$, with parents $p$ in the circuit, we can use the chain rule of differential calculus to write the derivatives as $\frac{\partial f}{\partial v} = \sum_p \frac{\partial f}{\partial p}\frac{\partial p}{\partial v}$. If $p$ is a multiplication node, $\frac{\partial p}{\partial v} = \frac{\partial (v \prod_{v'} v')}{\partial v} = \prod_{v'} v'$. Otherwise $p$ is a summation node, $\frac{\partial p}{\partial v} = \frac{\partial (v + \sum_{v'} v')}{\partial v} = 1$, where $v'$ are the other children of parent $p$. Note that $p$ can never be a variable or a constant because they are leaves in the arithmetic circuit. By sweeping down, $\partial f/\partial v$ can be computed for all nodes, since the derivatives $\partial f/\partial p$ for all its parents will have already been computed.

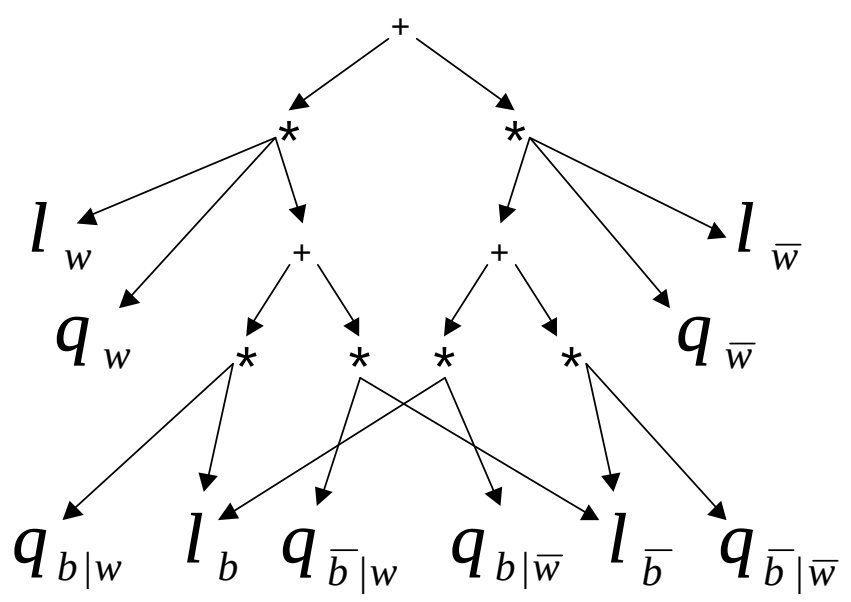

Figure 3: Arithmetic circuit for the belief network shown in Figure 1a.

To summarise, we recursively obtain the values $v$ in an upward pass and then recursively compute the derivatives in a downward pass. The upward and downward passes are also referred to as *sweeps*. For further details regarding arithmetic circuits, please see Darwiche (2003).

# 4 EVALUATING INFLUENCE DIAGRAMS WITH ARITHMETIC CIRCUITS

In this section, we present an algorithm to evaluate influence diagrams by converting them to arithmetic circuits.

**Normal Form Algorithm:** Given an influence diagram and evidence $\mathbf{e}$, determine the optimal strategy $s^*$ and maximal expected utility (MEU).

1. Convert the influence diagram to normal form, if it is not already in normal form. Let $S$ be the strategy variable and let $N_S$ be the number of possible strategies. Represent $S$ as an uncertainty and initialize $\theta_s = 1/N_S \forall\ s$ and any desired subset of evidence indicators for $\mathbf{e}$. Compile to obtain an arithmetic circuit.
2. Sweep up and down the circuit with evidence $\mathbf{e}' = \mathbf{e} \cup \{U = 1\}$.
3. The optimal strategy $s^* = \arg\max_s \frac{df}{d\lambda_s}(\mathbf{e}')$, and the optimal $MEU = \frac{\frac{\partial f}{\partial \lambda_{s^*}}(\mathbf{e}')\ N_S}{\frac{\partial f}{\partial \lambda_u}(\mathbf{e}') + \frac{\partial f}{\partial \lambda_{\bar{u}}}(\mathbf{e}')}$.

**Theorem 1.** *Every influence diagram can be evaluated with one upward and one downward sweep on an arithmetic circuit, using the normal form algorithm.*

*Proof.* Given an influence diagram in normal form, there is a single decision node with no parents. By treating that strategy as an uncertainty to be observed at its different values, the diagram can be represented by an arithmetic circuit. Following the approaches of Cooper (1988) and Shachter and Peot (1992), $P(U = 1|s, \mathbf{e}) = P(U = 1, s, e)/P(s, \mathbf{e})$ but $P(s, \mathbf{e}) = P(s)P(\mathbf{e})$ is constant by the initialization of $\theta_s$ and because $\mathbf{e}$ must be unresponsive to $s$. Therefore,

$$\begin{aligned} s^* &= \arg\max_s P(U = 1|s, e) \\ &= \arg\max_s P(U = 1, s, e) \\ &= \arg\max_s \frac{df}{d\lambda_s}(\mathbf{e}'), \end{aligned}$$

where the last step follows from Lemma 1. Similarly,

$$\begin{aligned} MEU &= P(U = 1|s^*, \mathbf{e}) \\ &= \frac{P(U = 1, s^*, \mathbf{e})}{P(s^*)P(\mathbf{e})} \\ &= \frac{\frac{\partial f}{\partial \lambda_{s^*}}(\mathbf{e}')N_S}{\frac{\partial f}{\partial \lambda_u}(\mathbf{e}') + \frac{\partial f}{\partial \lambda_{\bar{u}}}(\mathbf{e}')}, \end{aligned}$$

where the last step follows from Lemmas 1 and 2. □

Thus, any influence diagram can be evaluated in a single upward and downward sweep through an arithmetic circuit. The partial derivatives at the end of the second sweep allow us to compute the optimal policy and the MEU. The same arithmetic circuit can be used again with different evidence $\mathbf{e}$, which makes it easy to perform some value of information calculations. If we want to analyze multiple queries on the model at the optimal policy, then we can add $s^*$ to the evidence, and we can even compile the model with that observation. When the decision has been replaced by an optimal policy, we call the resulting diagram a *policy diagram* [Shachter and Kenley, 1989] and its circuit a *policy arithmetic circuit*.

Unfortunately, the number of strategies $N_S$ in a normal form influence diagram can be prohibitively large, since it assumes that any observations needed for one decision will be needed for all subsequent decisions. In experiments we ran on the influence diagram shown in Figure 1c, using the publicly available *ACE* system (http://reasoning.cs.ucla.edu/ace) with two possible values each for G, R, W, and B, the computed arithmetic circuit comprised 86 nodes and 134 edges.

There are alternative methods to convert influence diagrams into arithmetic circuits, some involving multiple compilations or the introduction of strategies for subsets of the decisions. We can imagine different influence diagrams for which one or another of these approaches would be better than the normal form algorithm. It seems even better in many cases, however, to maintain the original diagram in extensive form and modify the arithmetic circuit so that it can optimize over decisions, and we present these modified circuits in the next section.

# 5 DECISION CIRCUITS

Influence diagrams in extensive form can be evaluated using decision circuits. Along with the summation and multiplication nodes in arithmetic circuits, there are maximization nodes to make decisions. Decision circuits are similar to the modified arithmetic circuits introduced recently [Chan and Darwiche, 2006; Huang et al, 2006], except that the goal is to solve a decision problem.

## 5.1 Preliminaries

**Definition 2.** *A* <u>*decision circuit*</u> *is a rooted, directed acyclic graph whose leaf nodes are labelled with variables or constants and whose other nodes are either summation, multiplication, or maximization.*

In the same way that arithmetic circuits compute the value of a multi-linear function over the input variables, decision circuits compute the value of a backward dynamic programming function. The output, denoted by $g$, is computed at the root through an upward sweep, and partial derivatives are computed through a subsequent downward sweep. On the downward sweep, maximization nodes should be treated as if they were summation nodes.

The efficiency of decision circuit evaluation depends critically on the circuit choice, so it might be worthwhile to invest considerable time compiling a decision circuit even when it will be used only once. In the next section, we specify a particular method for constructing these circuits, but improved circuits are an open area of research.

## 5.2 Constructing decision circuits in the variable elimination order

Any valid influence diagram can be represented as a decision circuit. Decision circuits can be constructed in the variable elimination order with a bottom-up approach, based on the arithmetic circuit construction in Darwiche (2000). An alternative, top-down approach, based on junction trees, would be similar to the method in Park and Darwiche (2004), and would produce a similar decision circuit.

The factors (also known as potentials in the literature) associated with each variable are usually conditional probability tables. For our construction, our symbolic factors are the product of the evidence indicator and network parameter. Hence for a variable X and its parents $\mathbf{U}$, the factor $\phi(x, \mathbf{u}) = \lambda_x \theta_{x|\mathbf{u}}$.

Variable elimination for an influence diagram has some order restrictions not needed in variable elimination for a belief network. If $X$ is observed before decision $D$ then it should be eliminated after $D$, and otherwise it should be eliminated before $D$. Given any valid elimination order on the diagram, with the utility node being the first node to be eliminated, we can now construct a decision circuit bottom-up to perform the standard variable elimination operations, taking expectation for uncertainties and maximizing for decisions.

Let us demonstrate this by constructing a decision circuit for the example in Figure 1b. The factors for the three families are: $\phi(W)$, $\phi(B)$, and $\phi(UWB)$. The only valid elimination order is $U \prec W \prec B$ because $W$ is not observed before decision $B$ is made. If $g$ is the output for the circuit,

$$\begin{aligned} g &= \max_b \sum_w \sum_u \phi(w)\phi(b)\phi(w,b,u) \\ &= \max_b \phi(b) \sum_w \phi(w) \sum_u \phi(w,b,u) \\ &= \max_b \lambda_b \theta_b \sum_w \lambda_w \theta_w \sum_u \lambda_u \theta_{u|w,b} \end{aligned}$$

moving the summation inside and replacing the factors with the corresponding evidence indicators and network parameters. The decision circuit constructed from this process is shown in Figure 4.

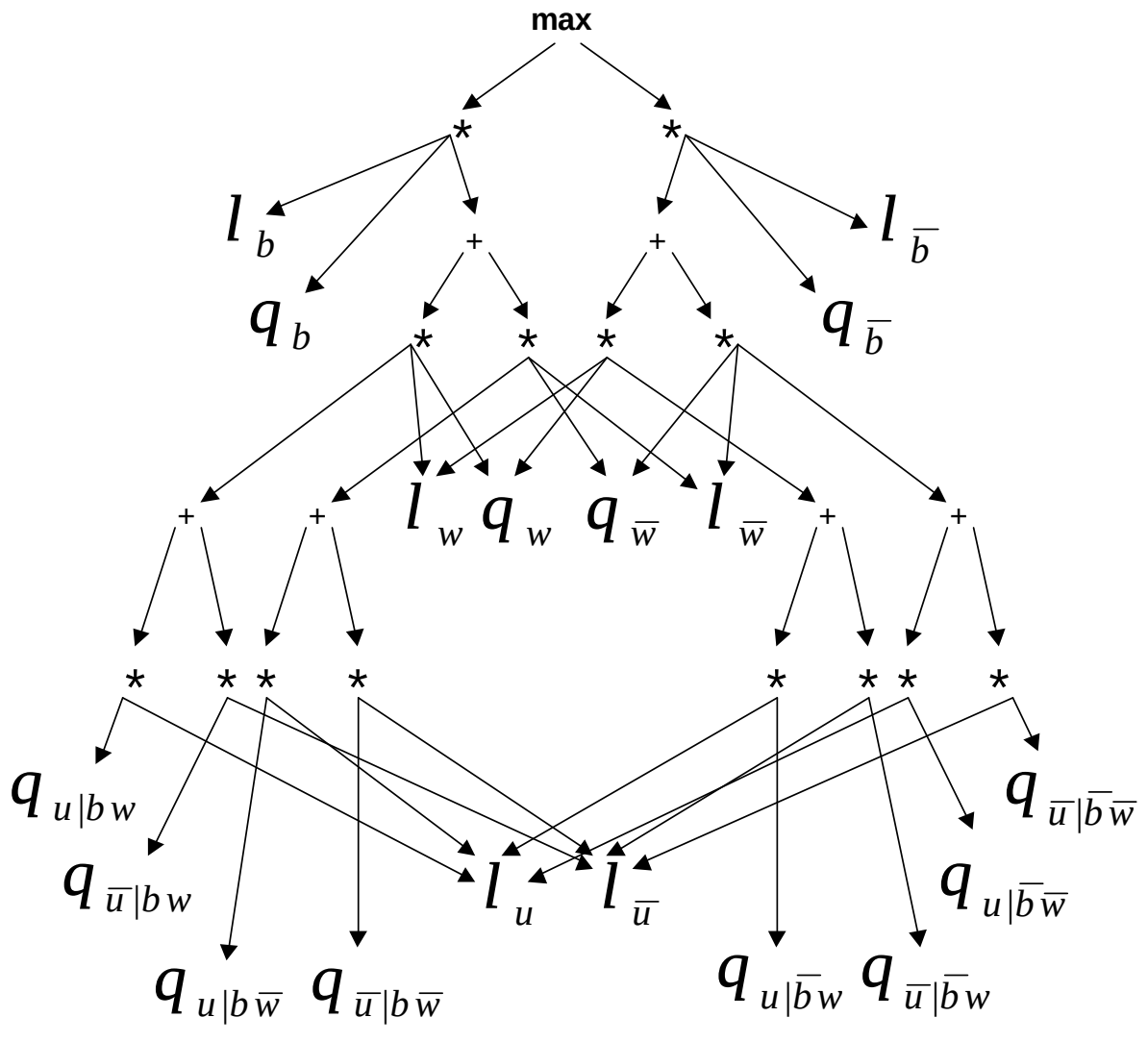

Figure 4: Decision circuit for the influence diagram shown in Figure 1b.

### 5.3 Evaluating influence diagrams with decision circuits

In this section, we present an algorithm using the constructed decision circuits to evaluate influence diagrams.

**Decision Circuit Algorithm:** Given an influence diagram and evidence $\mathbf{e}$, determine the optimal policies $\theta_{d|\mathbf{u}}$ and maximal expected utility (MEU).

1. Construct a decision circuit for the influence diagram, initializing $\lambda_d = 1$ and all $\theta_{d|\mathbf{u}} = 1$ for each decision alternative $d$.

2. Sweep up and down the circuit with evidence $\mathbf{e}' = \mathbf{e} \cup \{U = 1\}$. At each maximization node on the way up, choose the alternative $d^*$ with the highest value, breaking ties arbitrarily, and set $\theta_{d|\mathbf{u}} = 0$ for all other alternatives $d$.

3. The optimal $MEU = \frac{g^*(\mathbf{e}')}{\frac{\partial g}{\partial \lambda_u}(\mathbf{e}') + \frac{\partial g}{\partial \lambda_{\bar{u}}}(\mathbf{e}')}$.

For example, consider the circuit shown in Figure 4. Assuming that there is no evidence $\mathbf{e} = \varnothing$, we set all the evidence indicators to 1, except $\lambda_{\bar{u}}$, which is 0. We also initialize $\theta_b$ and $\theta_{\bar{b}}$ to 1, corresponding to the two alternatives. If $b$ is optimal, we set $\theta_{\bar{b}}$ to 0.

**Theorem 2.** *Every influence diagram can be evaluated with one upward and one downward sweep on a decision circuit constructed in variable elimination order, using the decision circuit algorithm.*

*Proof.* First, we want to show that each maximization node will select the best alternative unless the observation before that decision is inconsistent with the evidence $\mathbf{e}$, rendering the choice moot. Consider the first decision $D$ in the variable elimination order (i.e., $D$ is the decision that will be made latest), and let $d$ be any particular alternative for $D$. Partition the variables into three sets, $\mathbf{A}$, $\mathbf{B}$, and $\{U, D\}$, where $\mathbf{B} \cup \{U\}$ are the variables that are eliminated before $D$, and $\mathbf{A}$ are the variables eliminated after $D$. Let $\mathbf{a}$ be any particular instantiation of $\mathbf{A}$.

By the construction of the decision circuit, there is a maximization node for $D$ corresponding to observation $\mathbf{a}$, with each of its children corresponding to an alternative $d$. Let the value corresponding to that child be denoted by $v_d$. By the construction of the circuit, all of the variables in $\mathbf{B}$ are eliminated in the process of computing that value, so

$$v_d = \lambda_d \theta_{d|\mathbf{a}} P(U = 1, \mathbf{e_B}|d, \mathbf{a}) = P(U = 1, \mathbf{e_B}|d, \mathbf{a})$$

where $\mathbf{e_B}$ is the evidence involving the variables in $\mathbf{B}$. However, that evidence cannot be responsive to $D$, so $P(\mathbf{e_B}|d, \mathbf{a}) = P(\mathbf{e_B}|\mathbf{a})$. Therefore, unless that evidence is inconsistent with $\mathbf{a}$ making $P(\mathbf{e_B}|\mathbf{a}) = 0$, we obtain that

$$P(U = 1|d, \mathbf{a}, \mathbf{e_B}) = \frac{P(U = 1, \mathbf{e_B}|d, \mathbf{a})}{P(\mathbf{e_B}|\mathbf{a})} = \frac{v_d}{P(\mathbf{e_B}|\mathbf{a})}$$

Since all of these expressions are nonnegative, maximizing over $v_d$ corresponds to maximizing over $P(U = 1|d, \mathbf{a}, \mathbf{e_B})$, as desired, and the optimal policy is determined. If $v_d$ is maximal then the value computed at the maximization node is $v_d$. This process can then be performed on each subsequent decision in the elimination order.

The value computed at the root node will be $g^*(\mathbf{e}') = P(U = 1, \mathbf{e})$ at the optimal policies for all of the decisions. The MEU can then be computed as in Theorem 1. □

The decision circuit algorithm using a circuit constructed in variable elimination order can be analyzed in the same way as arithmetic circuits constructed in variable elimination order by Darwiche (2000). Key parameters are the number of nodes $n$ and the treewidth $w$ of the elimination order, which reflects the efficiency of that elimination order under the order restrictions. This makes it competitive with the best current algorithms for evaluating influence diagrams [Jensen et al, 1994].

**Lemma 4.** *A decision circuit constructed in variable elimination order has time and space complexity of the order of $O(n\,exp(w))$.*

**Theorem 3.** *Evaluating an influence diagram using a decision circuit has time complexity no worse than the order of $O(n\,exp(w))$.*

*Proof.* A sweep through the decision circuit is linear in the size of the circuit, and is thus $O(n\,exp(w))$ from the result on space complexity of the circuit in Lemma 4. Evaluating an influence diagram entails constructing a decision circuit, and performing two sweeps. Each of these three tasks is $O(n\,exp(w))$. Hence the entire process has time complexity of the order of $O(n\,exp(w))$. □

## 6 CONCLUSIONS AND FUTURE RESEARCH

Decision circuits can be used to evaluate and analyze influence diagrams efficiently, and they are a natural extension to arithmetic circuits. In this paper, we have shown their construction in variable elimination order, which does not fully exploit the asymmetry in most practical problems. Techniques using local structure could create even more compact circuits, and that would be especially valuable if the circuit were used to answer multiple queries or to respond in real time. This would create opportunities to address important problems for which exact solutions have remained intractable.

A significant advantage of arithmetic and decision circuits over traditional methods is that they provide a wealth of information for sensitivity analysis as a byproduct of the two sweeps. The partial derivatives and higher order derivatives answer many important sensitivity analysis queries [Darwiche, 2000; Darwiche, 2003; Chan and Darwiche, 2004]. In the case of decision problems, value of information analysis can be performed by evaluating the decision circuit at different evidence instantiations.

The algorithms in this paper could be enhanced by a simple preprocessing step now standard in the literature [Shachter, 1998; Nielsen and Jensen, 1999; Shachter, 1999]. Using the global structure of the influence diagram, we can determine a subset of the observations available at the time of the decision that are actually requisite or sufficient to be observed to make the best decision. This can significantly reduce the number of strategies or optimal policies to be computed, reducing the time and space needed. This approach could potentially be extended to the local structure as well, incorporating the asymmetry in most practical problems of the requisite observations needed to make decisions.

Another easy enhancement is a modeling feature that would allow the parents of a decision to be conditional as well as informational. If a particular decision alternative $d$ is not available for some instantiation of the observations at the time the decision is made, $\mathbf{u}$, this could be represented in the network parameters at initialization, setting $\theta_{d|\mathbf{u}} = 0$, much the same as analysis could be performed without a particular alternative $d$ by initializing its evidence indicator, setting $\lambda_d = 0$. This adds some significant modeling power at no computational cost.

Once the optimal policies have been determined using a decision circuit, or any particular strategy has been specified, we can transform the influence diagram into a policy diagram by treating the decision nodes as uncertainties and using the policies as conditional probabilities. This allows us to perform multiple queries under a particular policy. This analysis can be done in the decision circuit itself, but it may be even better to compile a policy arithmetic circuit, the arithmetic circuit for the policy diagram, instead. Since there are no order restrictions on the policy arithmetic circuit, state-of-the-art approaches could be used to construct the best possible circuit, exploiting the determinism introduced by the decision policies themselves. In that case, the decision circuit algorithm and the policy arithmetic circuit compilation could be performed offline, and queries on the compact policy circuit could be analyzed efficiently online.

**Acknowledgements**

This research was partially funded by the Global Climate Energy Project. We thank John Weyant for his support, the UCLA Automated Reasoning Group for the use of their software, and the anonymous reviewers for their feedback.